\documentclass[runningheads]{llncs}
\usepackage{graphicx}
\usepackage{tikz}
\usepackage{comment}
\usepackage{amsmath,amssymb} % define this before the line numbering.
\usepackage{color}
\usepackage[pagebackref=true,breaklinks=true,colorlinks,bookmarks=false]{hyperref}
\usepackage[accsupp]{axessibility}  % Improves PDF readability for those with disabilities.
\usepackage{pdfpages}
\newcommand{\squeezeup}{\vspace{-4mm}}
\newcommand{\squeezeupsmall}{\vspace{-2mm}}

\usepackage{xspace}
\usepackage{booktabs}
\makeatletter
\DeclareRobustCommand\onedot{\futurelet\@let@token\@onedot}
\def\@onedot{\ifx\@let@token.\else.\null\fi\xspace}

\def\eg{\emph{e.g}\onedot} 
\def\ie{\emph{i.e}\onedot}

\def\etal{\emph{et al}\onedot}

\newcommand*{\defeq}{\mathrel{\vcenter{\baselineskip0.5ex \lineskiplimit0pt
                     \hbox{\scriptsize.}\hbox{\scriptsize.}}}%
                     =}
\makeatother

\begin{document}
% \linenumbers
\pagestyle{headings}
\mainmatter
\def\ECCVSubNumber{5993}  % Insert your submission number here
\renewcommand{\baselinestretch}{0.9}

\title{RepMix: Representation Mixing for Robust Attribution of Synthesized Images} % Replace with your title

% %\begin{comment}
% \titlerunning{ECCV-22 submission ID \ECCVSubNumber} 
% \authorrunning{ECCV-22 submission ID \ECCVSubNumber} 
% \author{Anonymous ECCV submission}
% \institute{Paper ID \ECCVSubNumber}
% %\end{comment}
%******************

% \begin{comment}
\titlerunning{RepMix for Image Attribution}
\author{Tu Bui\inst{1}\orcidID{0000-0001-6622-9703} \and
Ning Yu\inst{2} \and
John Collomosse\inst{1,3}}
\authorrunning{T. Bui et al.}
\institute{University of Surrey
\email{t.v.bui@surrey.ac.uk} \and
Salesforce Research
\email{ning.yu@salesforce.com} \and
Adobe Research \email{collomos@adobe.com}
}
% \end{comment}
%******************
\maketitle

\begin{abstract}
Rapid advances in Generative Adversarial Networks (GANs) raise new challenges for {\em image attribution}; detecting whether an image is synthetic and, if so, determining which GAN architecture created it.  Uniquely, we present a solution to this task capable of 1) matching images invariant to their semantic content; 2) robust to benign transformations  (changes in quality, resolution, shape, etc.) commonly encountered as images are re-shared online. In order to formalize our research, a challenging benchmark, Attribution88, is collected for robust and practical image attribution. We then propose RepMix\footnote{This work was supported by EPSRC DECaDE Grant Ref EP/T022485/1.}, our GAN fingerprinting technique based on representation mixing and a novel loss. We validate its capability of tracing the provenance of GAN-generated images invariant to the semantic content of the image and also robust to perturbations. We show our approach improves significantly from existing GAN fingerprinting works on both semantic generalization and robustness. Data and code are available at \href{https://github.com/TuBui/image_attribution}{https://github.com/TuBui/image\_attribution}.

% We propose a robust and practical technique for GAN fingerprinting; detecting whether an image is synthetic and, if so, determining which Generative Adversarial Network (GAN) architecture created it.  Uniquely, we are able to trace the provenance of GAN-generated images invariant to the semantic content of the image. Our core contributions are a representation mixing and a novel loss to improve zero-shot generalization of the approach to unseen semantic classes at training time.  We consider the use case of images circulating online, and the need for users to detect and attribute synthetic images to GANs.
% We therefore train our model to encourage invariance to benign transformations  (changes in quality, resolution, shape, etc.) commonly encountered as images are re-shared.   We show our approach improves significantly on both semantic generalization and robustness of prior GAN fingerprinting work.

\end{abstract}

\section{Introduction}
% narrative: Synthesised images are getting better overtime with the like of Stylegan, to the extent that it can deceive human perception. While it has many creative applications, it also raises concerns over misinformation. It is important to distinguish synthesised images from real. Recent work has shown that synthesised images have distinct high frequency artifacts (cite stylegan2, DCT) that separate them from real images. Typical detection model could achieve near perfect real/fake classification accuracy. A harder problem is attributing synthesised images to its source models.   Recent work has already shown initial success at attribution of generative imagery, but has 2 limitations. First, the attribution is tied to specific training model and configurations -- a model trained at slightly different settings (even random seeds) could lead to degrade in attribution performance. Second, it is unclear how robust the attribution model is against visual perturbations during daily-life redistribution, where the subtle artifacts meaningful for tracing the image back to source could be easily altered. In this work, we propose a new challenging image attribution benchmark.  

Generative imagery is transforming creative practice through intuitive tools that enable controllable and high quality image synthesis. The photo-realism achievable by recent Generative Adversarial Networks (GANs) is often indistinguishable from real imagery \cite{nightingale2022ai}; it is difficult for a lay user to tell if an image is synthetic, or to tell images generated by one GAN from those generated by another. Yet, understanding the provenance of visual media has never been more important -- to help ensure creative rights, and to mitigate the spread of misinformation due to abuses of GAN technology. In the near future, parameterizable generative imagery may even begin to challenge or replace traditional stock photography.  Tools to trace an image to the GAN that created it are  urgently needed to ensure the authenticity and proper attribution of images shared online.

Recent work has already shown initial success at detecting synthetic imagery \cite{wang2020cvpr,beyondspec,fakespotter} and attribution of generative imagery \cite{yu2019,reverseeng2021,dct2020} (`GAN fingerprinting') to a GAN source. Particularly, Wang \etal \cite{wang2020cvpr} suggest that today GANs share some common technical flaws that could be easily distinguished from real images. However, image attribution is generally more challenging than synthesis detection due to the diversity in GAN classes; also it is inconclusive what sort of fingerprint a GAN model leaves in its output imagery. Existing image attribution methods, despite reporting near-saturated performance, have two setbacks. First, they mostly focus on attributing images to specific GAN models, which is impractical because a single change in training data, training metaparameters (\eg learning rate, optimizer, training iterations ...) or even random seed results in a different GAN model \cite{yu2019}. It would be more practical to attribute synthetic imagery to the underlining GAN architecture rather than specific GAN models. Second, the effects of perturbations on synthetic images are largely underestimated. Current works often experiment with few image transformations such as blurring, JPEG compression, random crop \cite{yu2019,dct2020,wang2020cvpr} which does not reflect the real-life perturbations that online imagery is subjected through redistribution. Such perturbations could deteriorate GAN fingerprint which is reported to lay between the medium and high frequency bands in an image \cite{zhang2019ifs}.       

The foremost contribution of this paper is a solid benchmark for image attribution, where a GAN class is represented by several GAN models trained on different semantic datasets, and images are subjected to various sources of perturbations. We then propose a novel method to robustly determine the fakeness of an image, and if so, which GAN architecture was used. Both our benchmark and proposed method address two key limitations of existing approaches:
% The contribution of this paper is a novel method for determining whether a GAN generated an image, and if so, which GAN architecture was used.  Recent work has already shown initial success at detecting synthetic imagery \cite{yu_class} and attribution of generative imagery \cite{yu_attr,dct} (`GAN fingerprinting') to a specific trained GAN model. Whilst we share a similar goal, our contribution addresses two key limitations of these existing methods:

{\bf 1. Semantic generalization.}  Existing GAN fingerprinting methods trained on images of a particular class of object (e.g. faces) typically fail on images of other object classes.  This is because prior works focus on attribution to one of several GAN models seen at training time. Uniquely, we address the new problem of attributing images of unseen semantic class to the GAN {\em architecture} that created them. In doing so, we formalize a new problem (attribution to GAN architecture rather than model), and propose a novel representation mix-up training strategy so as to equip GAN fingerprinting with semantic generalization over unseen models producing images containing unseen object classes.

{\bf 2. Robustness to benign transformation.}  Images often undergo non-editorial (benign) transformations, such as quality, resolution, or format change as they are redistributed online \cite{cai,nguyen2021oscar,black2021deep}.  Existing GAN fingerprinting techniques exploit artifacts in the GAN generated images in the pixel domain \cite{yu2019} or frequency domain \cite{dct2020} that are removed or corrupted via redistribution process, causing attribution to fail.  In some cases, GANs are actively trained to introduce such artifacts.  We employ a contrastive training strategy to enable our GAN attribution model to discriminate GAN architectures passively, based upon artifacts that are seldom removed via benign transformation upon images.

\section{Related work}
{\bf Generative Adversarial Networks (GANs)} \cite{gan} have shown outstanding performance in many downstream image synthesis tasks: photo-real blending and in-painting \cite{ganinpainting}, super-resolution \cite{gansr}, facial portrait generation \cite{ganportrait}, manipulation \cite{ganmanip}, and texture synthesis \cite{gantexture,yu2019texture}. GANs have been also applied to bridge multiple modalities such as geometry \cite{gangeometry}, audio \cite{ganaudio}, or sketch \cite{gansketch}. Our work focuses upon  unconditional GANs \cite{stylegan3,stylegan2,stylegan,progan,cramergan,mmdgan,sngan} to avoid introducing additional constraints when producing synthetic images.

{\bf Content provenance} explores the attribution of media to a trusted source (\eg a database or blockchain \cite{archangel1,archangel2}). Image provenance systems typically rely upon embedded metadata \cite{cai,origin}, watermarking \cite{hameed2006,devi2009,profrock2006,baba2009} or perceptual hashing \cite{dhn2016aaai,dsh2016cvpr,dpsh2016cjai,hashnet2017iccv} to perform visual search robust to the kinds of non-editorial transformation encountered online. Some methods are trained to fail in the presence of digital manipulation \cite{nguyen2021oscar}, whilst others are explicitly trained to match such content and highlight any manipulation \cite{black2021deep,blackcvmp} between the query and matched original. Regardless of applications, robustness and generalization are crucial for content provenance. This is usually addressed via data manipulation (augmentation, data mixing, adversarial attack), implicit representation learning (kernel methods, disentanglement) or explicit learning strategy (ensemble, meta-learning) \cite{wang2022generalizing}. In this aspect, RepMix can be considered as a blend of data manipulation (new data is created by mixing existing data points) and representation learning (mixing is performed at feature level). 

{\bf Digital forensics} methods detect and localize image manipulations in the `blind' \ie without a comparator.  The recent `deep fake detection challenge’ (DFDC) \cite{kaggledf} identified several approaches to detect GAN generated images or image regions,  either upon its statistical properties \cite{zhang2020manip,mantranet2019cvpr} or  current limitations of GAN methods (\eg human blinking \cite{blink}).  Our approach contributes most directly to this area, seeking to determine both the presence, and the source of, synthetic imagery. As such we are aligned with recent GAN fingerprinting work.  Prior work has explored this problem mainly for facial images, seeking to identify the model \cite{yu2019,dct2020,ding2021bmvc} or the architecture and metaparameters \cite{reverseeng2021}.  All these works are passive; the practicality of GAN identification is limited by reliance upon fragile signals within an image that are easily destroyed by benign transformation.  In order to mitigate this, Yu \etal instead propose to modify the GAN training to inject a robust fingerprint into the synthetic image \cite{yu2021artificial,yu2022responsible}.  However such approaches require active participation of the GAN creator, and all remain limited to images of a single semantic class.  Our fingerprinting approach is passive and robust to both unseen semantic classes and benign transformation, presenting a further step toward practical GAN attribution in the wild. %\ning{Cite and discuss the most recent ICCV'21 paper ``Towards Discovery and Attribution of Open-world GAN Generated Images''. Be inspired from their Related Work section.}

Most of the above approaches attribute images towards specific GAN models. Although Ding \etal \cite{ding2021bmvc} attempts to learn an architecture-specific attributor, their work only covers GAN models of different training seeds. Reverse Engineering \cite{reverseeng2021} shows that GAN architecture parameters could be traced even for unseen GAN models, however such fine-grain attributions mean each GAN class is represented by 1 GAN model; and the robustness of the model is still inconclusive. Recently, Girish \etal \cite{opengan2021} proposes to automatically discover a new GAN cluster for unseen synthesized images, at the cost of iterative evolution of the attributor. While we share a similar goal with \cite{opengan2021} in term of architecture-specific attribution, our work scope limits at a closed world problem (\ie attribution on a fixed set of GAN classes), instead we focus on the generalization on unseen semantic and transformations. 

\section{The Attribution88 benchmark}
\label{sec:data}
\begin{figure}[t!]
    \centering
    \includegraphics[width=1.0\linewidth]{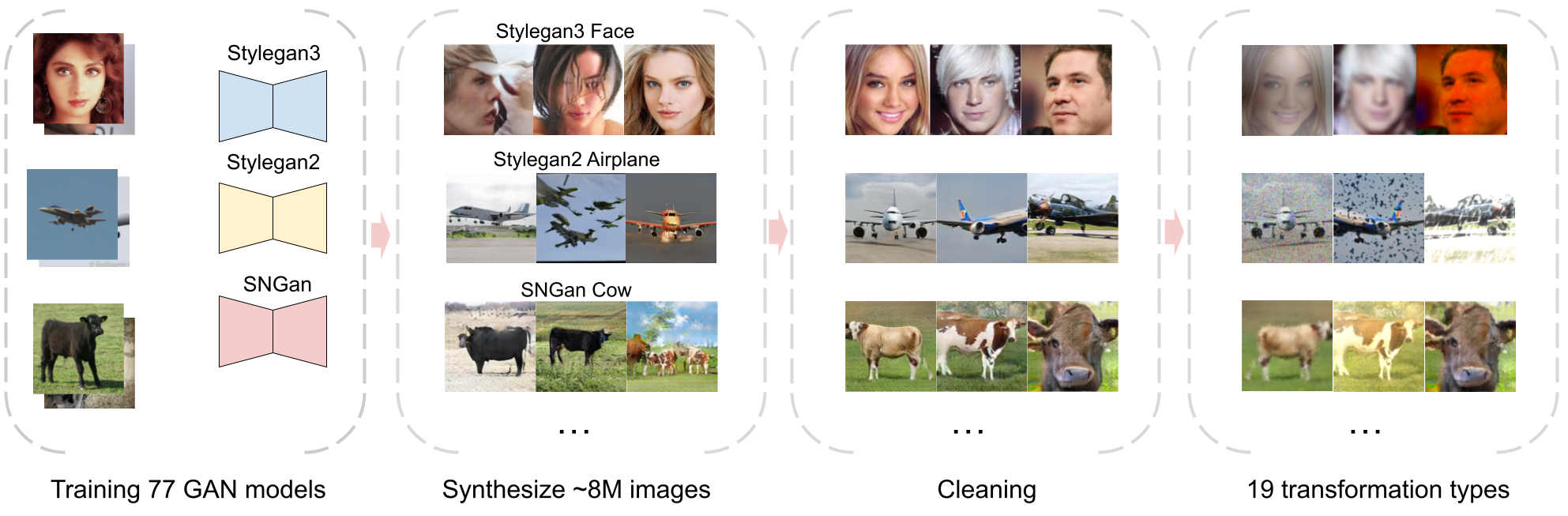}
    \caption{Illustrating the construction of Attribution88; a new dataset and benchmark that we contribute for synthetic image detection and attribution.  }
    \label{fig:data}
\end{figure}
% \begin{figure}[t!]
%     \centering
%     \includegraphics[width=1.0\linewidth]{figs/dataset.png}
%     \caption{Mean images and spectrums of several generator classes and semantics. For each inset, top: mean image before (left) and after (right) random perturbation, bottom: corresponding mean of DCT images.}
%     % \ning{Highlight the relation between this set of visualization and our insights/reasoning.}}
%     \label{fig:data}
% \end{figure}
The most popular attribution dataset in literature is introduced by Yu \etal \cite{yu2019}, containing 5 classes (Real + 4 GANs) of a single semantic object. Each GAN class is represented by one GAN model, thus the learned fingerprint could be entangled with semantic features. This is also not an absolute benchmark since only the GAN models are released (rather than the synthesized images) and there is not a fixed train/test split. Existing approaches \cite{yu2019,dct2020,reverseeng2021} report different results on this dataset, even for the common baselines. Additionally, the reported performance is near saturated. It is important to have a fixed and more challenging benchmark for image attribution. The new benchmark should have GAN classes tied to the GAN design/architecture rather than specific GAN models, meaning images from the same GAN class could come from different model training instances. While we could simply vary the training random seeds (\eg \cite{ding2021bmvc}) or other metaparameters to create different model instances of a same GAN, we leave the configuration of these parameters of each GAN model fixed to recommended settings for optimal generative quality. Instead, for each GAN class, we train multiple models on different sets of image objects (semantics). The new benchmark is more challenging as attribution must be agnostic to semantics.  %\ning{Highlight the existing benchmarks are not agnostic to semantics. Discussion the dataset differences from ``Towards Discovery and Attribution of Open-world GAN Generated Images''.}  

We introduce \textbf{Attribution88} - a new dataset made of 8 generator classes and 11 semantics (Fig.~\ref{fig:data}). We start with 5 generator classes (Real, Progan \cite{progan}, Cramergan \cite{cramergan}, Mmdgan \cite{mmdgan}, Sngan \cite{sngan}) as proposed in \cite{yu2019}, then add 3 most recent classes of the StyleGAN family (Stylegan \cite{stylegan}, Stylegan2 \cite{stylegan2} and Stylegan3 \cite{stylegan3}). For semantics, we choose 10 objects and scenes from the LSUN dataset \cite{lsun} plus the popular CelebA face dataset \cite{celeba}. We note that CelebA is structurally aligned and well curated as compared with other semantic sets, but it is widely used for image attribution/synthesis and adds  diversity to our benchmark. For each semantic set, we randomly select 100k images for training the 7 GAN models above, and a disjoint 12k images to serve as {\em real} images for the attribution task. We use pretrained GAN models when available, otherwise they are trained from scratch using public code, outputting 128$\times$128 images (more details in Sup.Mat). %\ning{``x'' should be replaced by ``$\times$'' everywhere.}

Next, we generate 100K images per GAN model, resulting in 7.7M synthesized images. Since some images have visible artifacts, we clean them to improve challenge and quality by first extracting  perceptual features (of synthesized and real images) using  InceptionV3 \cite{perceptualfeat}. We then use K-Means (k=100) to cluster the synthesized images, determine the closest real image for each, and sort the synthesized images according the distance to its closest real image. We then pick top-k (k=120) images in each group, assuming the images closest to a real one have the highest quality. This process helps retain a balance between diversity and realism of images.  Overall, we obtain 12K images for each of 8 generator sources (Real plus 7 GANs) and 11 semantics, totally $\sim$1M images. We further partition each set to 10K training, 1K validation and 1K test images. In our experiments, we expose only 6 semantics ({\em CelebA Face, Bedroom, Airplane, Classroom, Cow, Church Outdoor}) in training and evaluate on all test images (including 5 unseen semantic classes: {\em Bridge, Bus, Sheep, Kitchen, Cat}).  

\textbf{Perturbations}. Images circulated online are  subjected to benign perturbations, from mild transformations such as image resizing to strong ones like noises and enhancement effects. It is important to be robust against these. To this end, we employ ImageNet-C \cite{imagenetc}, a popular benchmark for evaluating classification robustness. ImageNet-C contains 19 common types of corruption, including various additive noises, blurring and effects, each has 5 different corruption levels. Similar to  \cite{imagenetc}, we only expose 15 transformations to training while the test set is subjected to all possible transformations. 

%The combination of unseen semantics and perturbations makes Attribution88 the most challenging benchmark for image attribution to date. Fig.~\ref{fig:data} shows examples of mean image in pixel and frequency domain for several generator sources and semantics, as well as the effect of perturbations on them. It can be seen that the characteristic  patterns of several GAN architectures are fade away or even removed under the presence of such perturbations, adding challenge to our benchmark. 

\section{Methodology}

\begin{figure}[t]
    \centering
    \includegraphics[width=0.9\linewidth,height=4cm]{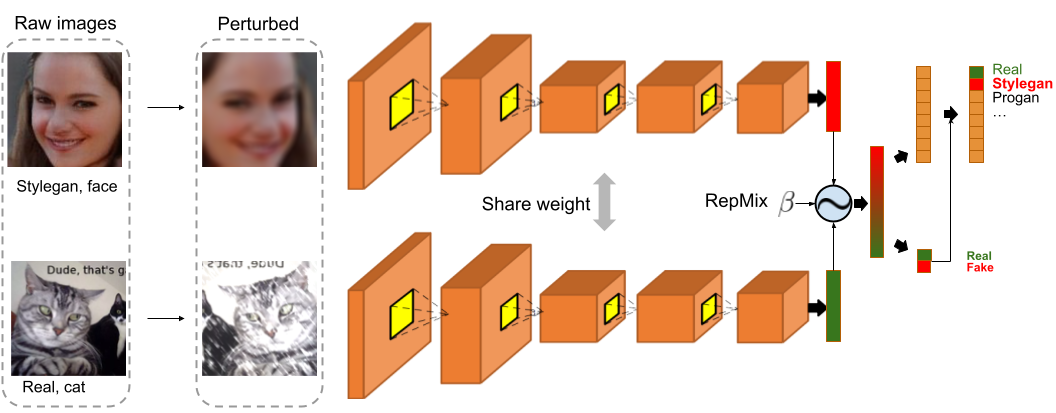}
    \caption{CNN architecture of our image attribution model. A pair of images is passed though the earlier layers of the CNN model, gets mixed in the RepMix layer before passing to  later layers. Training is regulated by a compound loss (see Sec.~\ref{sec:loss}).}
    \squeezeup
    \label{fig:arch}
\end{figure}

Synthetic image attribution is  a classification problem \cite{yu2019,dct2020,reverseeng2021,beyondspec}. In our case, the classes correspond to the GAN architectures from which the images are generated. Unlike semantic classification which relies on discriminative features of salient objects, the features useful for image attribution are often subtle and may deteriorate due to noise or other image perturbations \cite{yu2019,beyondspec}. In order to learn an attribution model robust against (even unseen) semantics and perturbations, we propose RepMix - a simple feature mixing mechanism to synthesize new data from interpolation between existing data points, then learn to predict the mixing ratio. Fig.~\ref{fig:arch} shows an overview of our approach. Our key technical advancements include (1) the RepMix layer that performs feature mixing between generator classes and (2) the compound loss to predict the mixing ratio for classification. %\ning{Please articulate how the prediction of mixing ratio can help for classification.}

\subsection{Representation Mixing (RepMix) Layer}
\label{sec:repmix}
% \ning{For each vector notation, people usually use a bold font. Please apply this principle to the other sections.}
Suppose we have a training set $\mathcal{X}=\{(\mathbf{x}_i, s_i, y_i), i=1,2,...\}$ where an image $\mathbf{x}_i$ has semantic label $s_i \in \mathcal{S}$ and source label $y_i \in \mathcal{Y}$ (which includes real and a set of GAN source labels). Our goal is to learn  mapping $\mathbf{x}_i$ to $y_i$ agnostic to $s_i$. 

Given a training image pair $\mathbf{x}_i$ and $\mathbf{x}_j$ which could either share or differ in source and semantic labels, we first project both images to an intermediate feature space using a nonlinear mapping function $f_e$: %\ning{Do $\mathbf{x}_i$ and $\mathbf{x}_j$ have to have the same semantics? The same source? Why or why not? I saw some descriptions in the next section but please repeat them here for clearance.}
\begin{equation}
    \mathbf{u}_i = f_e(\mathbf{x}_i);\qquad \mathbf{u}_j = f_e(\mathbf{x}_j)
\end{equation}
where $f_e(.)$ could be the earlier layers of a CNN module. The intermediate representations are input to our RepMix layer:
\begin{equation}
    \mathbf{u} = M_{\beta}(\mathbf{u}_i,\mathbf{u}_j) \defeq \alpha * \mathbf{u}_i + (1-\alpha)*\mathbf{u}_j 
\end{equation}
with random weight $\alpha$ generated from a certain distribution (here we draw $\alpha$ from a beta distribution\footnote{\url{https://en.wikipedia.org/wiki/Beta\_distribution}}, $\alpha \sim \text{Beta}(\beta, \beta)$).

Next, the mixed feature map $\mathbf{u}$ is projected into the output via a second mapping function (\eg the later layers of the CNN module):
\begin{equation}
    \mathbf{z} = f_l(\mathbf{u}) \in \mathbb{R}^D
\end{equation}
where D is the output dimension (D=256 in our work). We call $\mathbf{z}$ the embedding space as it directly precedes the objective function (subsec.~\ref{sec:loss}).% and contains the richest information needed for the image attribution task. %\ning{Why does $\mathbf{z}$ favor for robust attribution? I know it is inspired from the related work, but please spell the reasons out to make the writing more insightful.}

From an implementation perspective, RepMix is portable and can be inserted anywhere in any existing CNN architecture. Since it has no learnable parameters, it introduces minimal overhead at training time. And since it is used for training only, it can be removed during inference (equivalent to duplicating $\mathbf{x}_i$ to make $\mathbf{x}_j$ with the same semantic and source label). We consider RepMix an extension of MixUp and related work \cite{mixup,augmix,cutmix,deepaugmix} regarding the idea of mixing features. The difference is that existing work performs mixing in the raw image space, while RepMix performs at an intermediate layer. We argue that image attribution relies on subtle artifacts on an image (instead of salient objects) to distinguish real from fake as well as classifying different GAN sources. These useful artifacts could be overwritten or canceled out if images are mixed at pixel level, reducing overall performance (see Sec.~\ref{sec:exp}). 

\subsection{Compound loss}
\label{sec:loss}
To attribute an image to its source, existing works \cite{yu2019,dct2020,reverseeng2021} treat the class {\em real} the same way as other GAN classes prior to modeling classification with a cross-entropy loss. In fact, there is a hierarchical structure in our problem: an image can be either real or fake, if it is fake then it is synthesized from one of the GAN generators. Additionally, real images have a different distribution than GAN synthesized images (see sec.~\ref{sec:analyse}), therefore should be treated differently. To this end, we proposed a compound loss that takes into account real/fake detection and attribution at the same time.

We first detect the proportion of realness and fakeness scores in the mix up: %\ning{Text subscriptions in math notations should not be italic.}
\begin{align}
    z_\text{real} &= \mathbf{W}_\text{real}^T\mathbf{z}; \qquad z_\text{fake} = \mathbf{W}_\text{fake}^T\mathbf{z} \qquad \in \mathbb{R} \\
    \bar{z}_\text{real} &= \frac{{\rm e}^{z_\text{real}}}{{\rm e}^{z_\text{real}} + {\rm e}^{z_\text{fake}}}; \qquad \bar{z}_\text{fake} = \frac{{\rm e}^{z_\text{fake}}}{{\rm e}^{z_\text{real}} + {\rm e}^{z_\text{fake}}} \\
    L_\text{det} &= -\big(\alpha (1-y_i^*) + (1-\alpha)(1-y_j^*)\big)\text{log}(\bar{z}_\text{real})  \\
    &\qquad - \frac{1}{\left|\mathcal{Y}\right|-1} \big(\alpha y_i^* + (1-\alpha)y_j^*\big)\text{log}(\bar{z}_\text{fake})
\end{align}
where $\mathbf{W}_\text{real}, \mathbf{W}_\text{fake} \in \mathbb{R}^{D\times 1}$ are learnable filters, and pseudo label $y_i^*=0$ if $\mathbf{x}_i$ is real, otherwise 1 (same for $y_j^*$). This detection loss essentially measures the weighted cross entropy between real and fakeness of each image in the mix. Since there are generally more fake images than real in the training set, the fake term is scaled down by the number of GAN sources accordingly.    

The actual attribution task is performed via another cross-entropy loss, taking into account the real/fake-ness score:
\begin{align}
    \mathbf{z}_\text{attr} &= \mathbf{W}_\text{attr}^T\mathbf{z} + \mathbf{b} \in \mathbb{R}^{\left|\mathcal{Y}\right|} \\
    \hat{\mathbf{z}}_\text{attr} &= \begin{cases}
    z_\text{attr}^{(y_\text{real})} * \bar{z}_\text{real}\\
    z_\text{attr}^{(c)} * \bar{z}_\text{fake}\qquad \forall c \in \mathcal{Y} \backslash \{y_\text{real}\}
    \end{cases}\\
    L_\text{attr} &= -\alpha \text{log}(\frac{{\rm e}^{\hat{z}_\text{attr}^{(y_i)}}}{\sum_k{{\rm e}^{\hat{z}_\text{attr}^{(y_k)}}}}) - (1-\alpha)\text{log}(\frac{{\rm e}^{\hat{z}_\text{attr}^{(y_j)}}}{\sum_k{{\rm e}^{\hat{z}_\text{attr}^{(y_k)}}}})
\end{align}
where $\mathbf{W}_\text{attr} \in \mathbb{R}^{D\times \left|\mathcal{Y}\right|}$ and $\mathbf{b}$ are learnable weight and bias of a fully connected layer to linearly map our embedding $\mathbf{z}$ to the attribution logits. $(c)$ indicates the c-th element of the logit vector. Finally, the total loss is sum of the two above losses $L_\text{total} = L_\text{det} + L_\text{attr}$.

%Our compound loss is inspired by \cite{imagenet21k}, originally designed to handle classification with excessive number of classes. Uniquely, we integrate this loss to regularize representation mixing for image attribution.

\section{Experiments}
\label{sec:exp}

\subsection{Training details}
We use the Resnet50 architecture as the backbone for our RepMix model, with the final N-way classification layer replaced by a FC layer producing the 256-D latent code, followed by our compound loss (subsec.~\ref{sec:loss}). Our RepMix layer is inserted at the first FC layer for optimal performance (c.f. subsec.~\ref{sec:abl}), with $\beta=0.4$. Image pairs are randomly sampled from the training data, regardless of generator class and semantics. We do not enforce any constraint on sampling the image pairs to maximize all possible source/semantic combinations. During training we resize images to 256$\times$256 and augment with random crop to 224$\times$224, horizontal flip followed by a random {\em seen} ImageNet-C perturbation with activation probability of 95\%. We train our attribution models for maximum 30 epochs, with Adam optimizer and initial learning rate 1e-4, step decaying with $\gamma=0.85$ and early stopping based on validation accuracy. %We use the Pytorch library and a single GTX1080 GPU. 

\subsection{Baseline comparison}
\label{sec:baseline}
\begin{table}[t]
\caption{Performance of RepMix and other baselines on a control set that mimics Yu \etal \cite{yu2019} settings, and Attribution88 test set. Yu$^\dagger$ \etal refers to the implementation using the original public code}
\squeezeupsmall
\centering
\small
\resizebox{\linewidth}{!}{
\begin{tabular}{l|ccc|ccc}
\toprule
       & \multicolumn{3}{c|}{1 Sem., Clean}         & \multicolumn{3}{c}{Attribution88}  \\
       & Det. Acc. $\Uparrow$ & Attr. Acc. $\Uparrow$ & Attr. NMI $\Uparrow$ & Det. Acc. $\Uparrow$ & Attr. Acc. $\Uparrow$ & Attr. NMI $\Uparrow$ \\
       \midrule 
RepMix & \textbf{1.0000} & \textbf{0.9994} & \textbf{0.9975} & \textbf{0.9745} & \textbf{0.8207} & \textbf{0.6679} \\
Yu \etal \cite{yu2019} (reimp.) & 0.9910 & 0.9838 & 0.9458 & 0.9306 & 0.6784 & 0.4666   \\
Yu$^\dagger$ \etal \cite{yu2019} & 0.9888 & 0.9844 & 0.9455 & 0.9190 & 0.6322 & 0.4028   \\
DCT-CNN \cite{dct2020} & 0.9922 & 0.9838 & 0.9526 & 0.9001 & 0.6447 & 0.4061   \\
Reverse Eng. \cite{reverseeng2021} & 0.9976 & 0.9960 & 0.9834 & 0.8665 & 0.5637 & 0.3653 \\
EigenFace \cite{eigenface} & 0.8262 & 0.6538 & 0.4515 & 0.7829 & 0.1515 & 0.0034     \\
% Beyond Spectrum \cite{beyondspec} & & & & 0.8144 & 0.1245 & 0.0002 \\
PRNU \cite{prnu2019} & 0.8544 & 0.8482 & 0.7389 & 0.7845 & 0.1252 & 0.0003     \\
% Random & 0.781 & 0.125 & 0
\bottomrule
\end{tabular}}
% \ning{Highlight the best in bold and indicate each metric the higher/lower the better.}}
\label{tab:main}
\squeezeup
\end{table}

\begin{figure}[t!]
    \centering
    \includegraphics[width=0.9\linewidth,height=5.5cm]{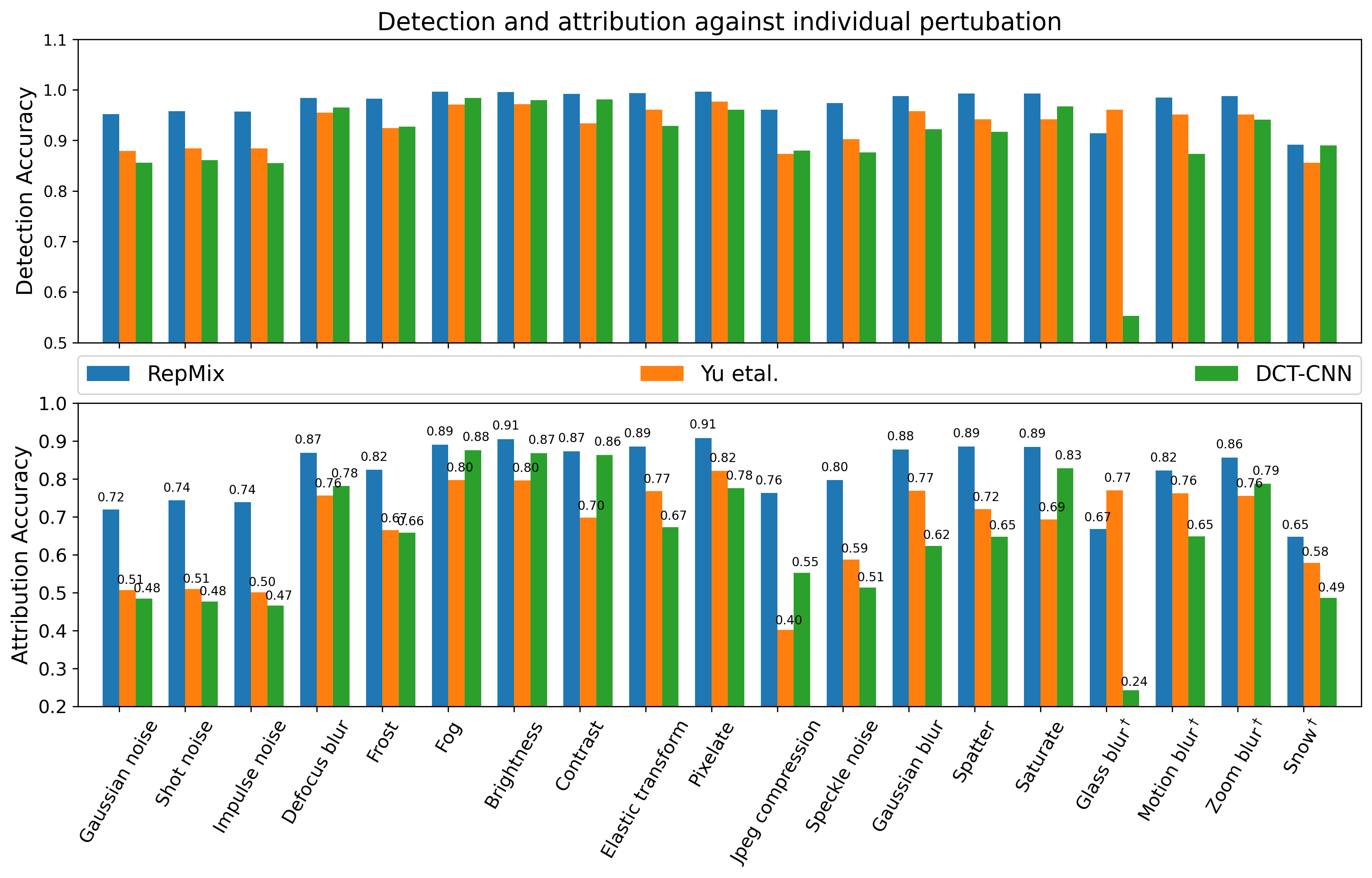}
    \vspace{-4pt}
    \caption{Detection and attribution performance of our proposed RepMix method vs. two baselines \cite{yu2019,dct2020} in the presence of different benign perturbations of the image.}
    \label{fig:aug_level}
    \squeezeup
\end{figure}
We compare our method with 5 baselines: (i) Yu \etal \cite{yu2019} attributes images via a simple fingerprinting CNN model; (ii) DCT-CNN \cite{dct2020} classifies images in the frequency space; (iii) Reverse Engineering \cite{reverseeng2021} models GAN architecture details such as number of layers and loss types to assist attribution; (iv) EigenFace \cite{eigenface} builds an Eigen model for each class and classify an image based on its maximum correlation with each model; (v) PRNU \cite{prnu2019} is similar to EigenFace but works on noise fingerprints of each class instead. The baseline models are trained using public code with the same data augmentation techniques as in the proposed method. We also provide our re-implementation of Yu \etal's approach. More details on the baseline implementation are in the Sup.Mat.  

To validate our training of the baselines and the GAN models, we also perform comparison on a replica of Yu \etal \cite{yu2019} dataset, denoted as {\em 1 Sem., Clean}. Specifically, we adopt their data cleaning method, use 5 classes (1 real and 4 GANs) as stated in \cite{yu2019} and without any ImageNet-C perturbation. The only difference is that we use our trained GAN models and we apply random crop and horizontal flip as the minimal augmentation during training and test. 

\textbf{Evaluation metrics.}
We report standard classification accuracy and Normalized Mutual Information (NMI) score \cite{opengan2021} that measures the dependence between the prediction and the target. Since {\em real} is one of the target classes, we are also interested in an auxiliary metric, detection accuracy, which is the proportion of images being correctly classified as {\em real} or {\em not-real}.

Tab.~\ref{tab:main} compares the performance of RepMix against baselines. The performance on the control set is comparable with existing work \cite{reverseeng2021,dct2020,yu2019}, with near-saturated accuracy on the deep learning approaches. Reverse Engineering is the highest scored baseline, next is DCT-CNN \cite{dct2020} which performs slightly better than Yu \etal \cite{yu2019}. RepMix achieves perfect detection accuracy and the best attribution accuracy and NMI. However, the baselines underperform on Attribution88. The frequency-based methods (DCT-CNN, Reverse Engineering) under-perform the pixel-based ones (Yu \etal). The complexity of our benchmark also causes the shallow methods to either fail completely (PRNU \cite{prnu2019}) or just  above random prediction (EigenFace \cite{eigenface}). We attribute these changes to the diversity of data (including unseen semantics) and severity of the perturbations. RepMix performs with 4\% and 14\% higher accuracy than the closest baseline on the detection and attribution scores. %This validates the efficacy of our representation mixing mechanism in robustifying the model learning (by enforcing a linear interpolation between input and output), improving invariance to semantic class (via mixing features from different semantics) and reducing the memorization effect common to neural networks \cite{bengio2017memorize}.  %\ning{Reason why.}

\subsection{Robustness against individual perturbation}
\label{sec:pertub}
To analyze the effects of individual perturbation on attribution performance, we evaluate RepMix and the closest competitors, Yu \etal \cite{yu2019} and DCT-CNN \cite{dct2020} on Attribution88 with  ImageNet-C perturbations applied on test images (Fig.~\ref{fig:aug_level}). JPEG compression and additive noise hinders the performance most significantly, especially on the two baselines, while other perturbation sources that transform blocks of neighboring pixels but do not replace them (\eg blurring) have less severe effects. DCT-CNN is particularly vulnerable to glass blurring. Performance on seen and unseen perturbations is comparable, indicating generalization of our models when being exposed to a large enough sources of augmentations during training. Additionally, detection performance is more robust than attribution, with detection standard deviation of 2.8\% across all perturbations versus 8.0\% attribution for RepMix (3.7\% vs. 12.1\% for Yu \etal method; 9.3\% vs. 16.8\% for DCT-CNN). 

\begin{table}[t]
\caption{Attribution errors caused by adversarial attacks on the Attribution88 test set at different levels of max perturbation $\epsilon$. Lower is better}
\squeezeupsmall
\centering
\small
\begin{tabular}{lcccccc}
\toprule
 Methods & $\epsilon=2/255$ & $\epsilon=4/255$ & $\epsilon=8/255$ &
 $\epsilon=16/255$ & $\epsilon=24/255$ & $\epsilon=32/255$\\
 \midrule
 RepMix & 0.1509 & 0.1952 & 0.2454 & 0.3008 & 0.3333 & 0.3572 \\
 Yu \etal \cite{yu2019} & 0.2113 & 0.2709 & 0.3328 & 0.3945 & 0.4303 & 0.4534 \\
 DCT-CNN \cite{dct2020}& 0.1545 & 0.2190 & 0.2831 & 0.3375 & 0.3642 & 0.3812\\
 \bottomrule
\end{tabular}

\label{tab:adv}
\squeezeupsmall
\end{table}

\subsection{Generalization on semantic and perturbation}
\label{sec:sem}
\begin{figure}[t!]
    \centering
    \includegraphics[height=3.5cm]{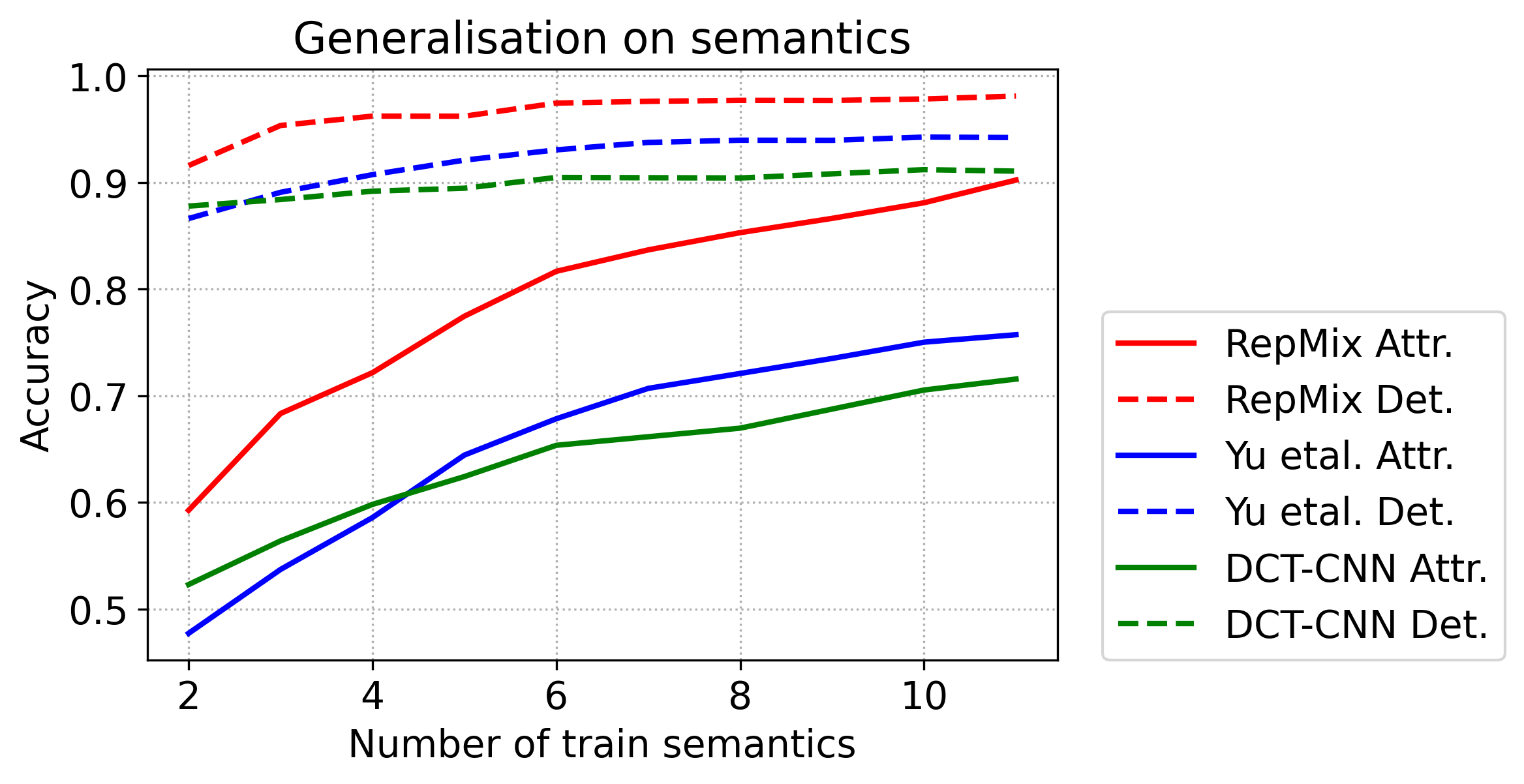}
    \includegraphics[height=3.5cm]{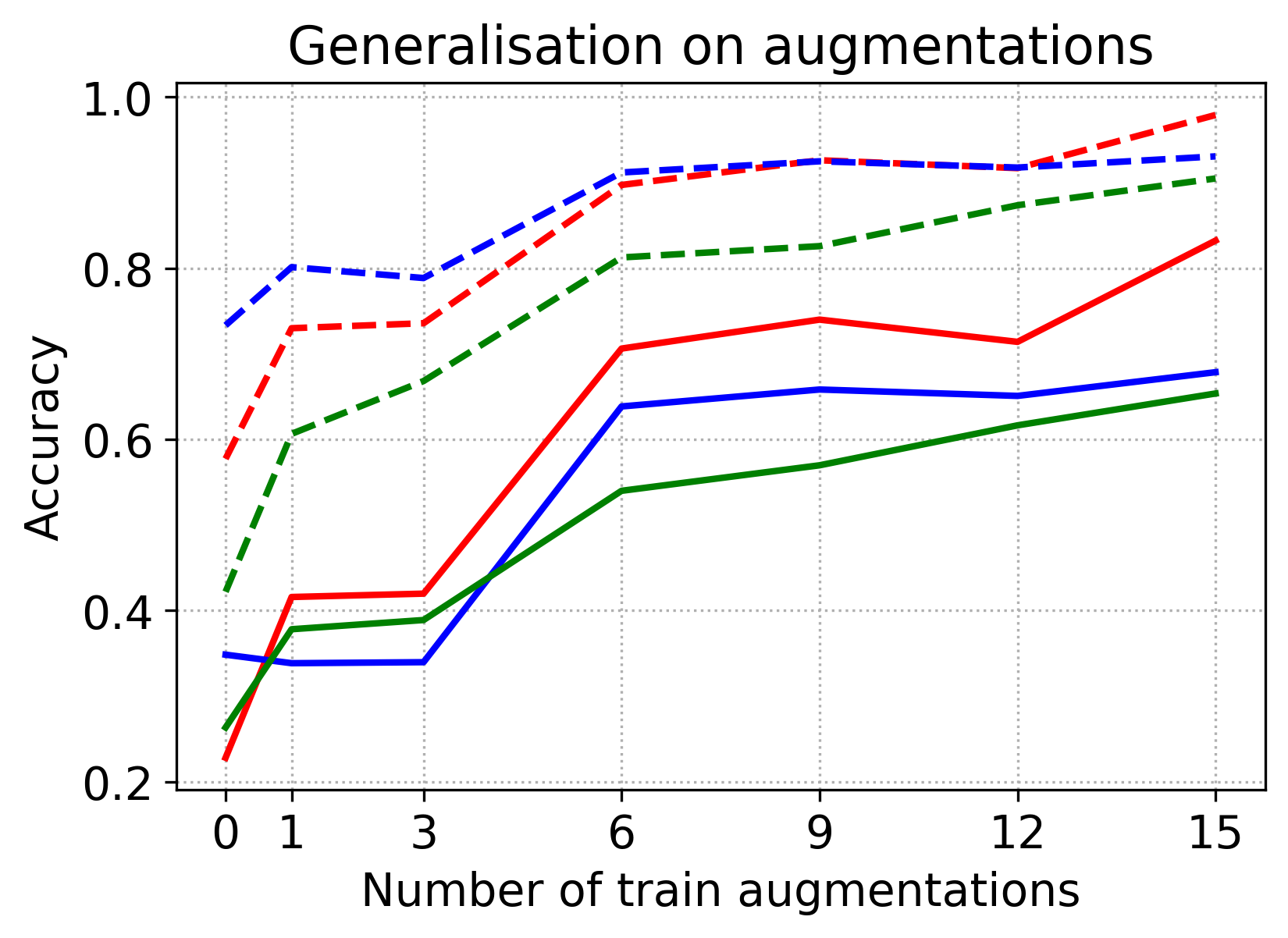}
    \caption{RepMix performance versus number of (left) semantics and (right) augmentations seen during training.}
    \label{fig:sem}
    \squeezeup
\end{figure}

We evaluate the generalization properties of RepMix, Yu \etal and DCT-CNN approaches under the circumstance of limited training data and data augmentation. Fig.~\ref{fig:sem} (left) depicts detection and attribution performance when the models are exposed to increasing number of semantics during training. We evaluate on the full Attribution88 test set. All 3 detection curves stabilize quite early with RepMix consistently maintaining a 3\% gap above other two methods. On attribution performance, the more training data leads to more rewarding results, with RepMix having better generalization capability, scoring from 59\% accuracy at 2 seen semantics to 90\% when all 11 semantics are exposed during training.

Fig.~\ref{fig:sem} (right) shows a similar trend as the number of data augmentation methods increases. We fix the number of training semantics at 6, and increase the number of augmentation methods from 0 to 15, and test on a held-out test set of 4 unseen perturbations. The overall trend is a boost in performance when exposing the models to more perturbations during training, with RepMix gaining more generalization power beyond 15 perturbations.

\subsection{Robustness against adversarial attacks}
\label{sec:adv}

Adversarial attacks introduce to an image a subtle layer of noise which is invisible to the naked eye but enough to change the prediction results of a model. Adversarial attacks work by diverting the gradient w.r.t input image toward the most plausible class other than the groundtruth. Repmix enforces a linear inter-class interpolation in the intermediate feature space, therefore is robust to adversarial attacks by design. To verify this, we perform untargeted whitebox attacks on Repmix, Yu \etal and DCT-CNN models using the I-FGSM method ~\cite{ifgsm}. We use 20 iterations of I-FGSM for every image in the Attribution88 test set and stochastic gradient ascend for optimization. Tab.~\ref{tab:adv} shows the attribution errors, which is the difference in attribution accuracy before and after adversarial attacks, at different noise levels. Although all methods suffer a performance drop and the severity is higher at higher noise tolerant levels (\ie $\epsilon$), RepMix is more robust than the other two approaches. At max perturbation $\epsilon=32/255$, RepMix accuracy is 2x higher than Yu \etal and DCT-CNN (46.35\% vs. 22.49\% for Yu \etal, and 26.34\% for DCT-CNN). Interestingly, DCT-CNN \cite{dct2020} has better resistance than Yu \etal \cite{yu2019}, probably because an images in frequency spectrum are visually more monotonous and alike than in the pixel domain thus would require more efforts (aka. iterations) from I-FGSM for a successful attack. 

\subsection{Ablation Study}
\label{sec:abl}
\begin{table}[t!]
\caption{Ablation study of RepMix exploring performance at attribution and detection whilst removing different design components, and alternate backbone choices}%\ning{Highlight the best in bold and indicate each metric the higher/lower the better.}}
\squeezeupsmall
\small
\resizebox{\linewidth}{!}{
\begin{tabular}{lccc}
\toprule
                     & Detection Acc. $\Uparrow$ & Attribution Acc. $\Uparrow$ & Attribution NMI $\Uparrow$ \\
\midrule 
All                  &  \textbf{0.9426} & \textbf{0.7400} & \textbf{0.5546}  \\
 w/o compound loss    & 0.9364 & 0.7204 & 0.5280   \\
 w/o RepMix            & 0.9296 & 0.7188 & 0.5205   \\
 w/o RepMix+Compound loss & 0.9283 & 0.7129 & 0.5167 \\
 w/o augmentation     & 0.7044 & 0.2762 & 0.0856  \\
\midrule
\multicolumn{4}{l}{Different backbones}                           \\
\midrule
VGG16             & 0.9493 & 0.7150 & 0.5315 \\ 
AlexNet           & 0.8818 & 0.5280 & 0.2817 \\ 
% Resnet101         & 0.9523 & 0.7430 & 0.5546 \\
% Densenet121       & 0.9504 & 0.7474 & 0.5673 \\
\bottomrule
\end{tabular}}

\squeezeup
\label{tab:ablation}
\end{table}

Tab.~\ref{tab:ablation} shows the performance of RepMix when removing one or several of its components or changing the backbone architecture. Without loss of generality we train and test our ablated models on a subset of Attribution88, with all 8 source classes but 2 semantics during training, and test on 4 semantics (2 seen and 2 unseen). Removing either RepMix layer or compound loss or both results in a drop in performance of all metrics. It can be seen that the compound loss does not benefit only the {\em real} class (small drop in detection accuracy when removing it), but the whole attribution (2\% drop). Finally, removing all ImageNet-C perturbations (leave only random crop and horizontal flip as the data augmentation method) significantly decreases the performance, even causes misleading real/fake detection (detection accuracy below random guess). We also replace Resnet50 with AlexNet \cite{alexnet} and VGG16 \cite{vgg}. AlexNet leads to a significant performance drop, with NMI score reduced by a half. VGG16 has comparable detection accuracy, but 2.5\% lower attribution score. More backbone experiments can be found on Sup.Mat.

\begin{figure}[t!]
    \centering
    \small
    \begin{tabular}{cc}
        \includegraphics[width=0.45\linewidth,height=3.7cm]{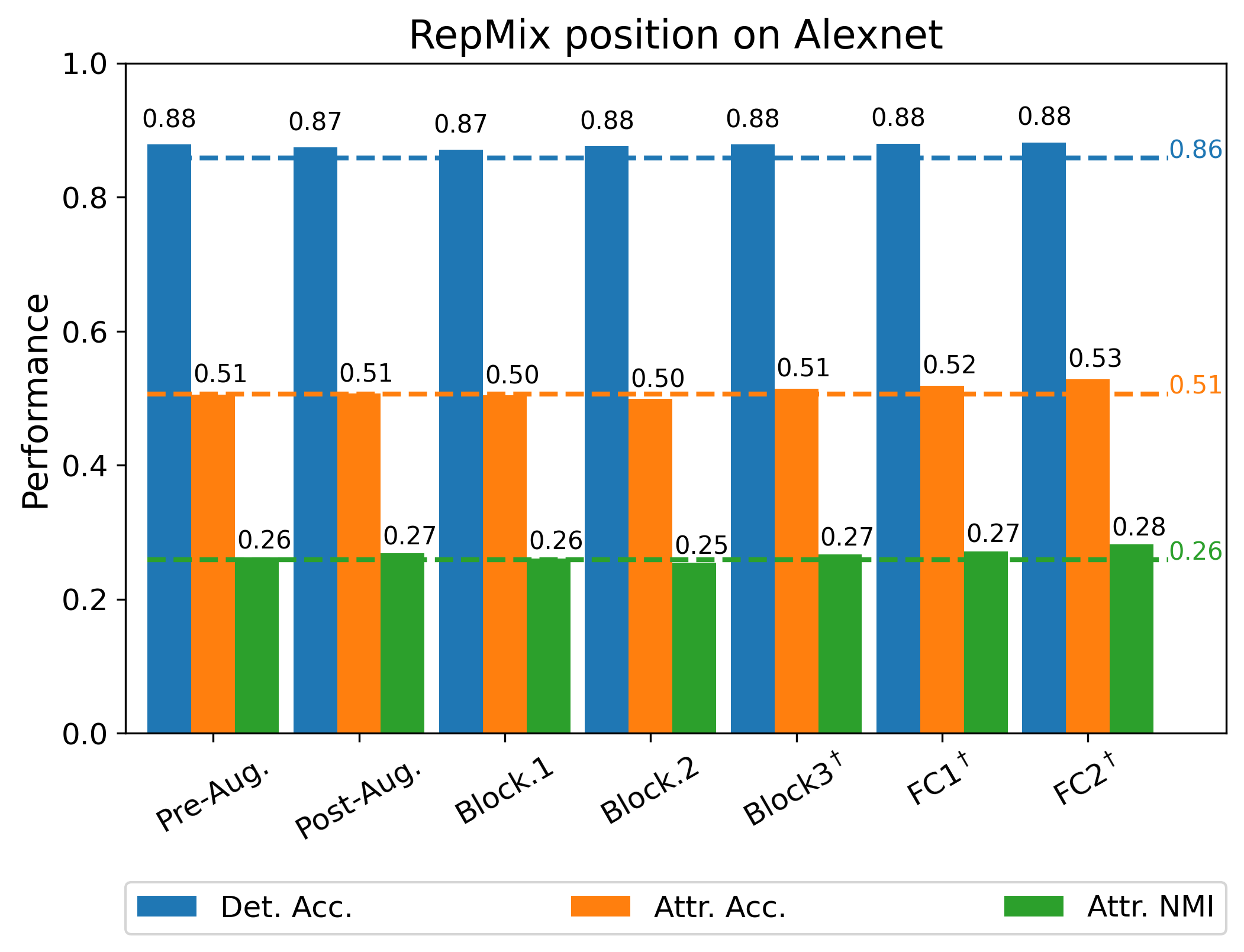} &
        \includegraphics[width=0.45\linewidth,height=3.7cm]{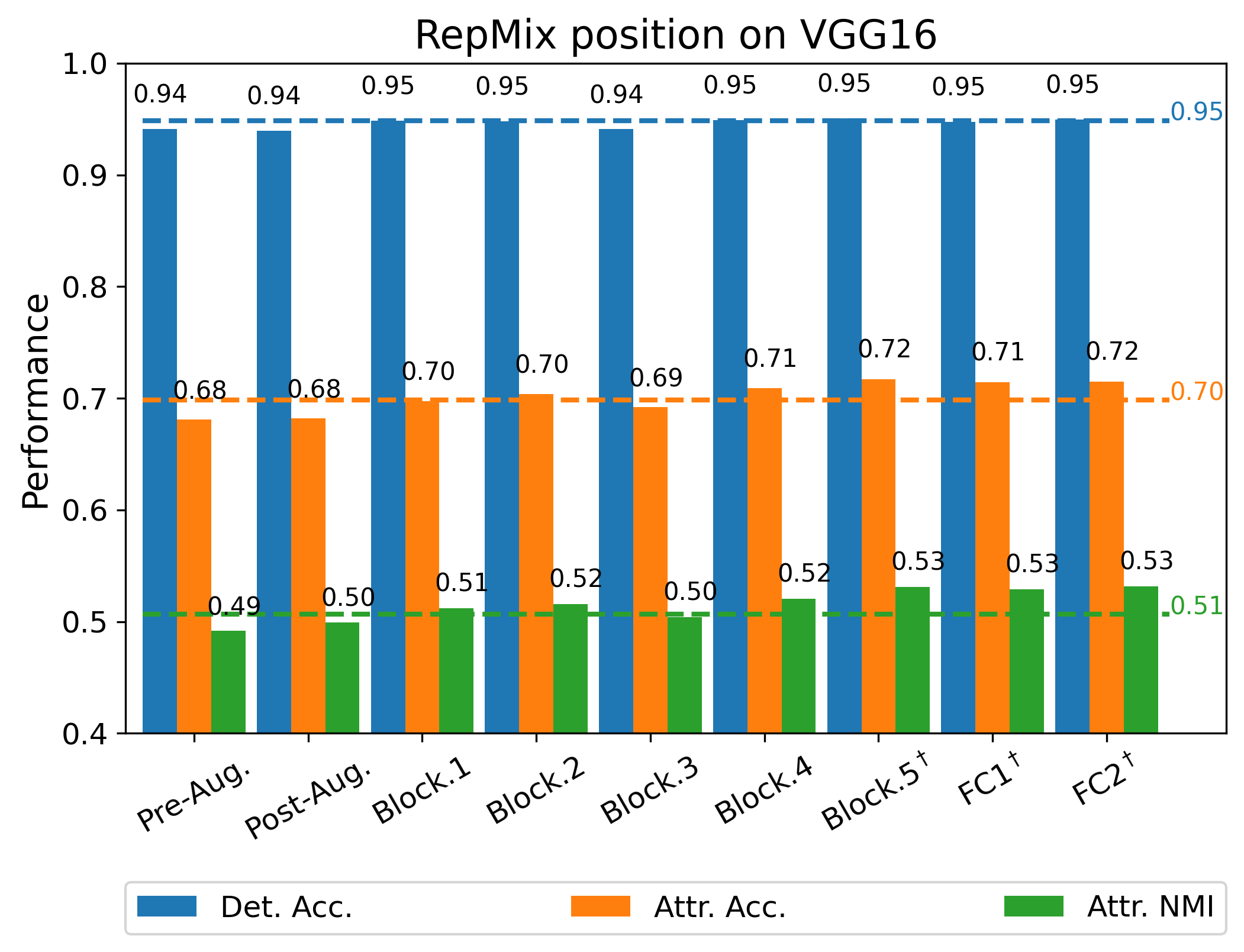} \\
        (a) AlexNet & (b) VGG16 \\
        \includegraphics[width=0.45\linewidth,height=3.7cm]{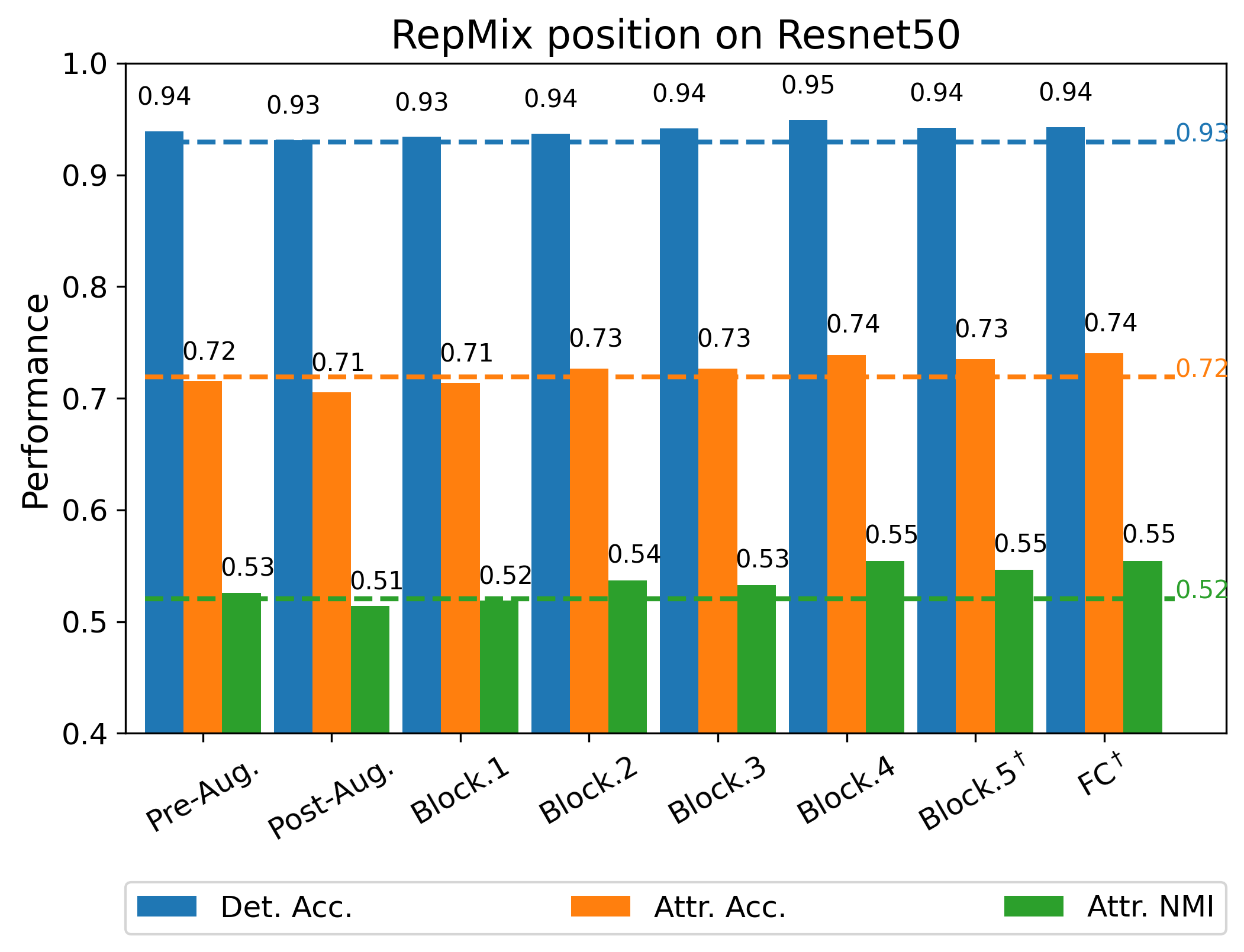} &  
        \includegraphics[width=0.45\linewidth,height=3.7cm]{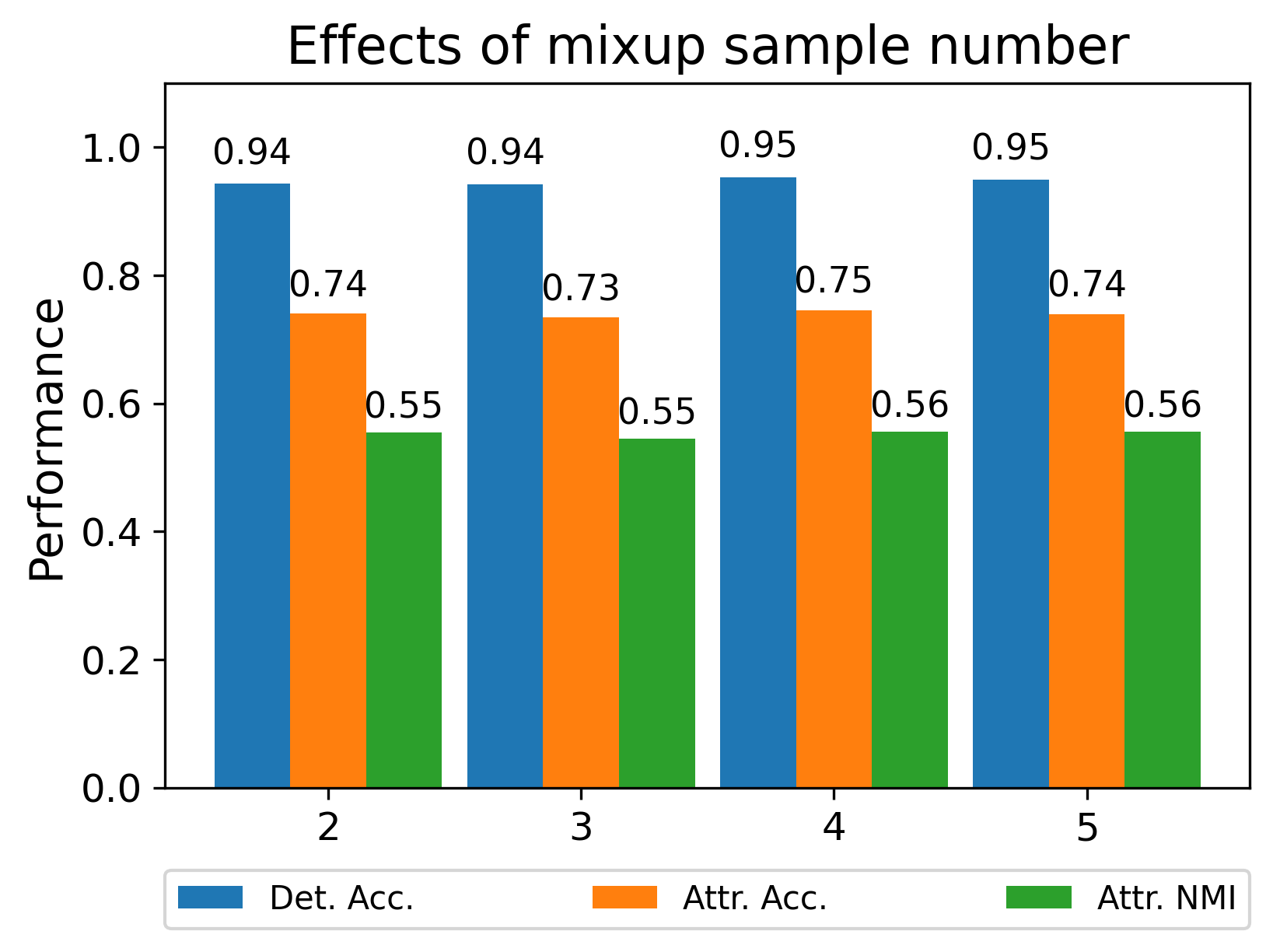} \\
        (c) Resnet50 & (d) Mixup samples on Resnet50
    \end{tabular}
    \caption{Effect of RepMix on different layers of (a) AlexNet, (b) VGG16 and (c) Resnet50. Dashed lines refer to baselines without mixing. $\dagger$ indicates the mixing is performed on 1-D feature map (either after Global Average Pooling or FC layer). (d) - The number of mix-up samples have marginal effect on performance of Resnet50.}
    \label{fig:abl_layer}
    \squeezeup
\end{figure}

\textbf{RepMix position}. We experiment with different positions of the RepMix layer in Resnet50, VGG16, and AlexNet. RepMix can be applied to input images at pixel level (equivalent to MixUp \cite{mixup}), before data augmentation (Pre-Aug.) or after it (Post-Aug.). Within the CNN layers, we insert RepMix after every pooling or FC layer. Fig.~\ref{fig:abl_layer} shows a similar trend across the three networks. Mixing images at pixel level does not improve performance; meaningful subtle artifacts are lost. Post-Aug mixing has the worst score since the image is exposed to double corruption. RepMix is more beneficial at the later layers of the networks, benefiting less on 2D feature maps and more on global representation (FC features). This can be seen from Fig.~\ref{fig:gradcam}, where the attention heatmap covers larger areas. In Fig.~\ref{fig:tsne}, semantic clusters appear even at the embedding layer. However, the GAN classification loss ensures  semantic features are weaker at the later layers while the GAN class signal is stronger. Thus, mixing representations at later layers is more beneficial.    

\textbf{Number of mixup samples}. We test with increasing number of samples to be mixed in RepMix layers. The beta distribution now becomes the Dirichlet distribution to accommodate more than two samples in a mixing group. Fig.~\ref{fig:abl_layer} (d) shows that increasing number of mixing samples has marginal boost in performance, with 1\% improvement at 4 mixing samples at most. 
% \begin{figure}
%     \centering
%     \includegraphics[width=0.5\linewidth]{figs/abl_samples.png}
%     \caption{Effect of number of mixup samples.}
%     \label{fig:abl_sample}
% \end{figure}

\subsection{Further analysis}
\label{sec:analyse}

\begin{figure}[t!]
    \centering
    \includegraphics[width=1.0\linewidth,height=6.5cm]{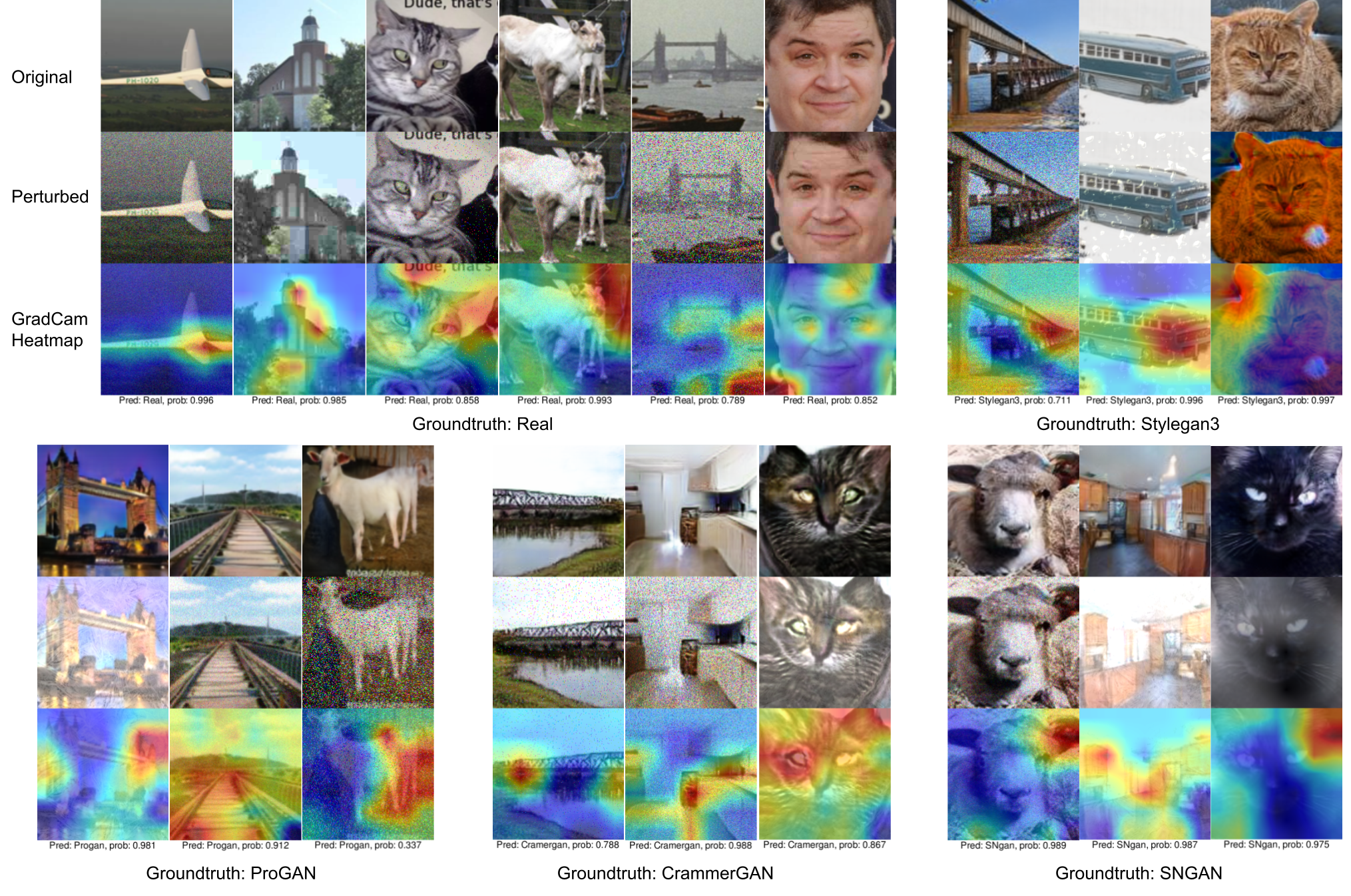}
    \caption{GradCAM visualization on unseen-semantic test images showing the visual artifacts contributing most signficantly to the GAN classification decision.}
    \label{fig:gradcam}
    \squeezeup
\end{figure}

\begin{figure}[t!]
    \centering
    \includegraphics[width=1.0\linewidth]{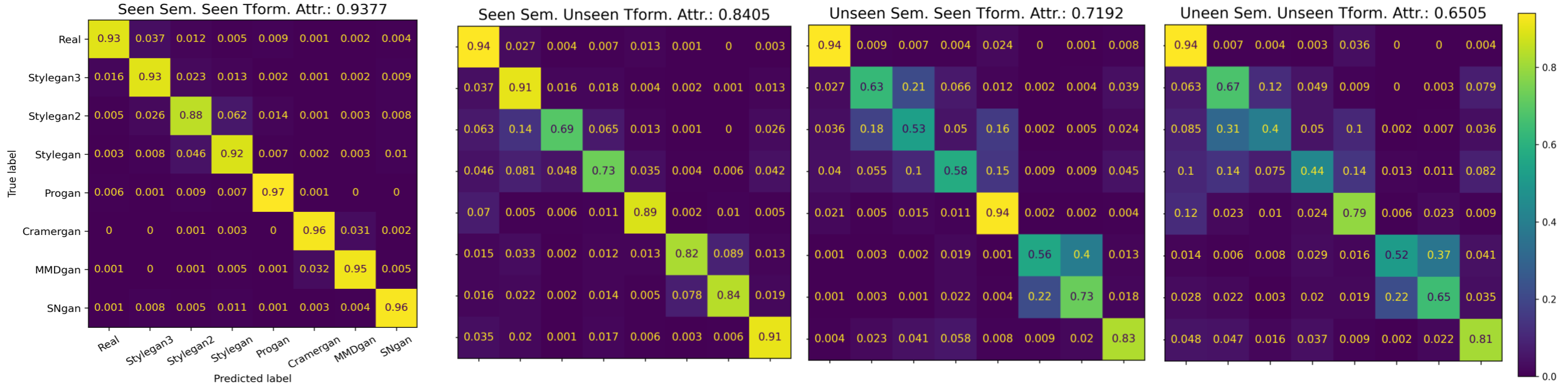}
    \caption{Confusion matrix of RepMix on seen/unseen semantic classes and on seen/unseen classes of image transformation applied to the test images.}
    \label{fig:conf}
    \squeezeup
    \squeezeupsmall
\end{figure}

\textbf{Real versus other classes}. We observe that the detection of real images is fairly robust to training data and perturbations and across various ablation settings (c.f. Sec.~\ref{sec:baseline}-\ref{sec:abl}. This interesting behavior is further demonstrated in Fig.~\ref{fig:conf}, where class real has the highest score and also appears the most consistent across the seen/unseen semantics and perturbations. 

To understand this behavior, we visualize the image regions that contribute the most to the prediction of our model using GradCAM \cite{gradcam}. Fig.~\ref{fig:gradcam} shows examples of GradCAM heatmaps for several images of {\em real} and other GAN classes, from both seen and unseen semantics as well as perturbations. For GAN classes, the heatmaps tend to highlight the edge regions which are often more resilient to perturbation attacks. For real images, GradCAM heatmap also focuses on background objects. We therefore reason that real images have a different distribution from synthesized images particularly because they have vivid background, which often attracts the attention of our attribution model.  

\begin{figure}[t!]
    \centering
    \includegraphics[width=0.9\linewidth]{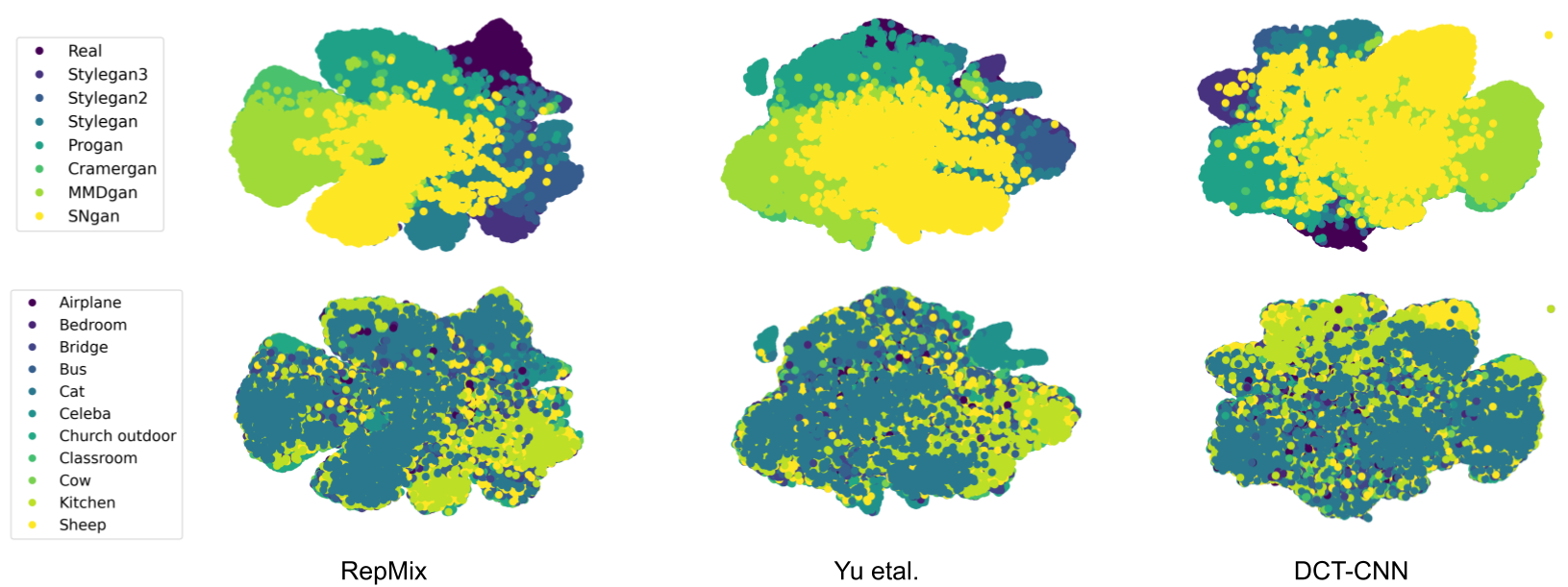}
    \caption{t-SNE visualization of Attribution88 test set using features extracted from RepMix (left) or Yu \etal (middle) and DCT-CNN approach.}
    \squeezeup
    \label{fig:tsne}
\end{figure}

\textbf{t-SNE visualization}. We visualize the embedding space $\mathbf{z}$ of RepMix computed on the Attribution88 test set using t-SNE \cite{tsne} 2D projection, and compare it with Yu \etal approach. Fig. \ref{fig:tsne} shows RepMix has better class separation and semantic fusion than Yu \etal. Nevertheless, both approaches have a mixed region in the middle of the t-SNE plots where classes are not well separated, which illustrates the challenge of the Attribution88 benchmark.

\begin{figure}[t!]
    \centering
    \includegraphics[width=0.9\linewidth,height=3.8cm]{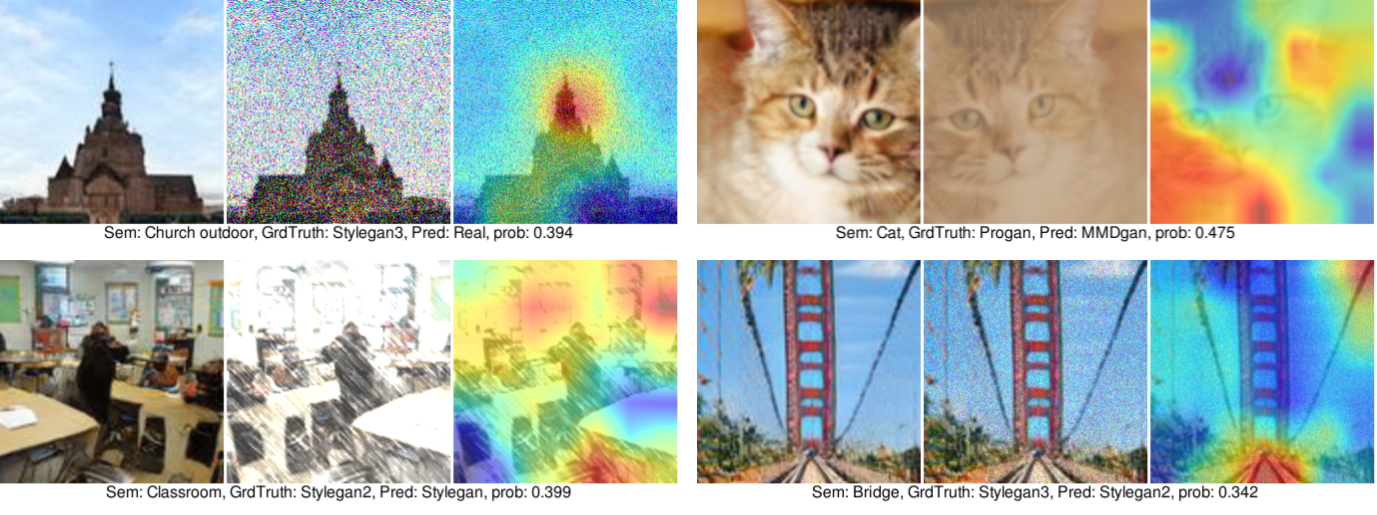}
    \caption{Examples of attribution failure. For each inset, left: raw image, middle: image after perturbation, right: GradCAM heatmap justifying its (wrong) prediction.}
    \label{fig:fail}
    \squeezeup
\end{figure}

\textbf{Limitations}. Fig.~\ref{fig:fail} shows examples where RepMix fails, often due to excessive perturbation that distort finer details of an image,  narrowing the gap between real/synthesis and between different GAN classes. Another  case shown is mis-classification between the three StyleGAN due to architectural similarity.

\squeezeup
\section{Conclusion}
\squeezeupsmall
We introduce a challenging image attribution benchmark, Attribution88, for detecting and tracing images to the originating GAN architecture, rather than the GAN model. We present a novel GAN fingerprinting technique that introduces strong zero-shot generalization to unseen semantic classes and unseen transformations, in contrast to prior work that generalizes poorly beyond a single class (\eg faces) even if trained with sight of those classes \cite{yu2019,dct2020}. We demonstrate detection accuracy of 97\% and attribution accuracy of 82\% on this new benchmark, without introducing any change to the GAN training process (per \cite{yu2022responsible}). Our method is particularly robust to detecting real images, by exploiting an unique feature that current GAN methods have not been able to fabricated. Future work could scale our experiments to even broader classes of GAN including conditional GAN frameworks, although we do not believe such experiments necessary to demonstrate the value of  benchmark or contrastive training and mix-up strategy in enabling class generalization for GAN attribution.

% \section{Acknowledgment}

% We have presented a novel GAN fingerprinting technique, capable of detecting and attributing synthetic images to the originating GAN model.  Our proposed fingerprinting architecture adapts DCT de-noising network to identify a characteristic (model-specific) signal in the frequency domain, robust to any benign (non-editorial) change to the image such as resolution, or re-compression/quality change.  We showed that training the model using a semantic mix-up strategy introduces strong zero-shot generalization to unseen semantic class, in contrast to prior work that generalizes poorly beyond a single class (\eg faces) even if trained with sight of those classes \cite{ningwork}.  We demonstrated classification accuracies of X\% and X\% respectively over seen and unseen classes for a bank of X pre-trained GAN models i.e. without introducing any change to the GAN training process (per \cite{scalable}).  Future work would….

% Future work: adversarial attacks on attribution models.

\clearpage
\bibliographystyle{splncs04}
\bibliography{egbib}



\section*{Supplementary material}

# RepMix: Representation Mixing for Robust Attribution of Synthesized Images [Supplementary Materials]

Tu Bui[1], Ning Yu[2], and John Collomosse[1,3]

[1] University of Surrey `t.v.bui@surrey.ac.uk`
[2] Salesforce Research `ning.yu@salesforce.com`
[3] Adobe Research `collomos@adobe.com`

## 1 More details of the Attribution88 benchmark

### 1.1 Training GAN models

To compose the Attribution88 benchmark, we need to train 77 GAN models for 7 GAN architectures and 11 semantics. Tab. 1 shows the source repositories that we employed to train those GAN models. We use the recommended settings. Specifically, Stylegan3 [7] has configuration stylegan3-t, batch size 32, and 20M training iterations. Stylegan2 [9] has configuration config-f, batch size 32, and 10M training iterations. Stylegan [8] settings are the same as Stylegan2 except using configuration config-a. Progan [6] has batch size 32 and 12M training iterations. Cramergan [2] has a g-resnet5 backbone, 256 dof-dim, gradient penalty of 10.0 and 150k training iterations. MMDgan [3] also has a g-resnet5 backbone, 16 dof-dim, gradient penalty of 1.0 and 150k training iterations. SNgan [11] has resnet backbone, batch size 32, hinge loss, and 100k training iterations. To avoid mode collapse, we checkpoint regularly and manually examine the quality and diversity of the output sampled images.

Despite our training efforts, several models are hard to train. We therefore use publicly released models if available. Specifically, there are pretrained Progan [6] models[4] for all the 10 LSUN semantics, which are used in Attribution88. For CelebA face where there does not exist an official pretrained Progan model, we use an unofficial version on tensor-hub[5]. For computational efficiency and eco-friendly validation, all synthesized images are downscaled to 128×128 if its public-release corresponding models were trained for higher resolutions. The diversity in our GAN model collection makes Attribution88 more practical and challenging. Example images are shown in Fig. 1.

---

[4] https://drive.google.com/drive/folders/15hvzxt_XxuokSmj0uO4xxMTMWVc0cIMU
[5] https://tfhub.dev/google/progan-128/1

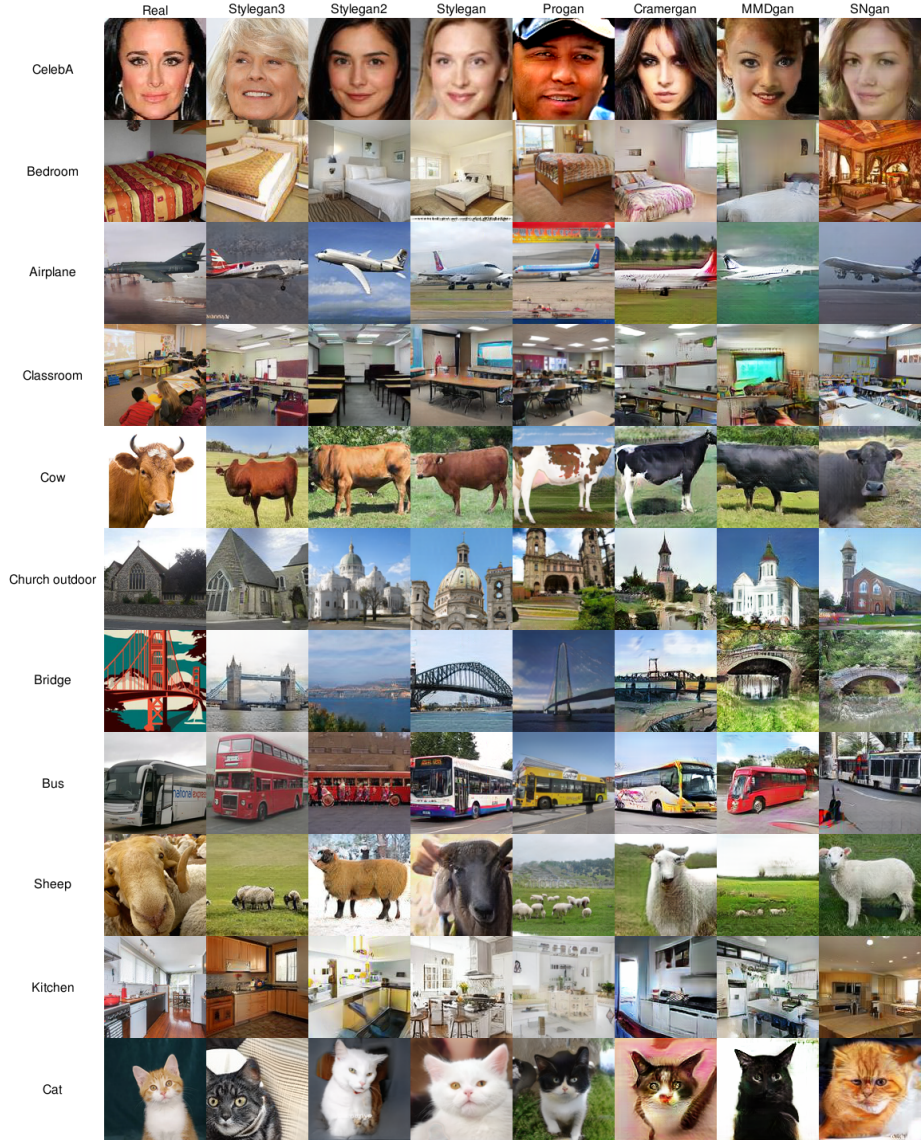

**Fig. 1.** Image examples of each source/semantic categories in Attribution88 benchmark.

### 1.2   Data cleaning

We describe in Sec.3 of the main paper our cleaning procedure for Attribution88. The goal is to remove images with artifacts without reducing the level of diversity in the generated data. Removing images with artifacts is achieved by selecting the synthesized images that are perceptually closest to a real one, while diversity

**Table 1.** Code repositories used in our GAN model training for Attribution88 benchmark

| Architecture | Codebase |
| --- | --- |
| Stylegan3 | https://github.com/NVlabs/stylegan3 |
| Stylegan2, Stylegan | https://github.com/NVlabs/stylegan2 |
| Progan | https://github.com/tkarras/progressive_growing_of_gans |
| Cramergan, MMDgan | https://github.com/mbinkowski/MMD-GAN |
| SNgan | https://github.com/pfnet-research/sngan_projection |

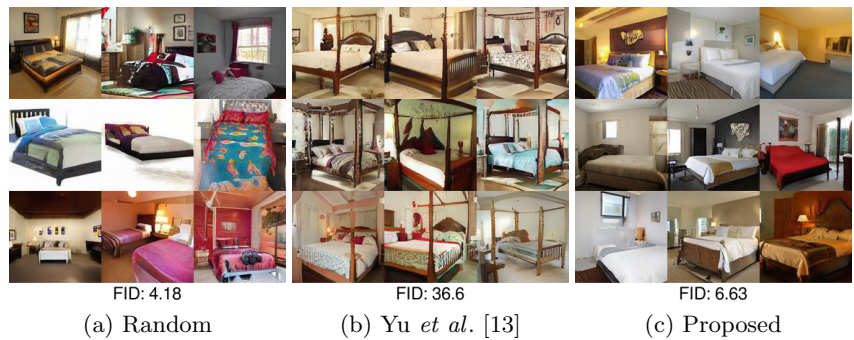

(a) Random    (b) Yu *et al*. [13]    (c) Proposed

**Fig. 2.** Data cleaning examples for Stylegan2 model source, LSUN Bedroom semantics. (a) random images without cleaning - high diversity but often contain artifacts; (b) data cleaning using Yu *et al*. method - high quality but lack diversity; (c) proposed method - balancing diversity and quality.

is maintained via clustering. We note that Yu *et al*. [13] also performed data cleaning, however their method sacrifices diversity for quality, leading to many similar images. Fig. 2 compares our data cleaning method with Yu *et al*. on Stylegan2 bedroom images. It is worth noting that, although models from the Stylegan family [8, 9, 7] have an internal truncation mechanism to enhance the overall quality of the generated images, artifacts are still present in some of them (Fig. 2a). Our cleaning method filters out the images with artifacts at a neglectable cost of only 1.5 degradation in FID, as opposed to 8.8x degradation in Yu *et al*. approach.

### 1.3 Perturbations

Fig. 3 shows the effect of 19 ImageNet-C perturbations [5] on an example synthesized image. These transformations further complicate the attribution task as it deteriorates fine-level features on the image. Fig. 4 illustrates the mean pixel and mean spectrum of several sources and semantics, before and after random perturbations. Unique GAN patterns are visually observable on the mean spectrum of the clean images, which are successfully exploited for image attribution

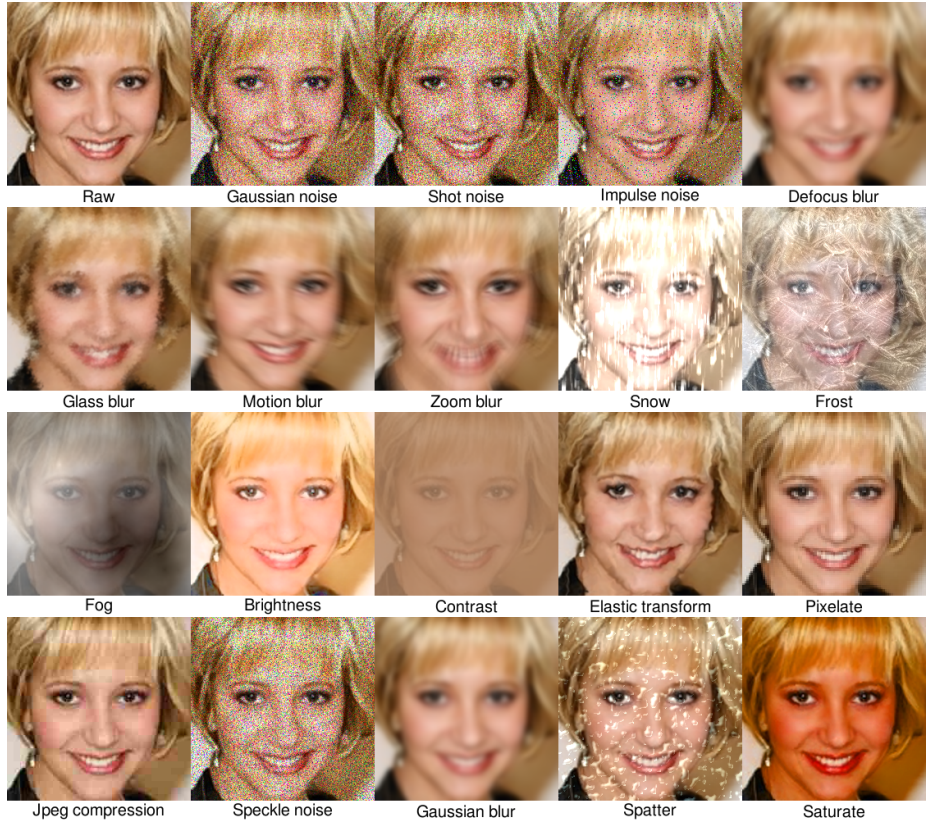

**Fig. 3.** Effects of different perturbations on a Stylegan3 CelebA generated image. All perturbations have strength 3 (out of 5). Best view in color when zoom in.

in DCT-CNN [4]. However, random perturbations make it difficult to recognize and attribute these patterns to the GAN architectures.

**CelebA versus the other semantics**. Fig. 4 also shows the difference between CelebA faces and the other semantics in Attribution88 benchmark. The visible mean pixel images regardless of generator sources demonstrate that face images are more aligned than the other semantic objects. This demonstrates the needs of validating an image attribution algorithm (and even an image synthesis algorithm) on multiple semantic sets, which is a design feature of Attribution88. The inclusion of CelebA along with other LSUN semantics also adds diversity to our benchmark.

## 2  Baseline implementation

Tab. 2 shows the code repositories that we employed to train and evaluate the baseline models. We follow the recommended settings and only change the data

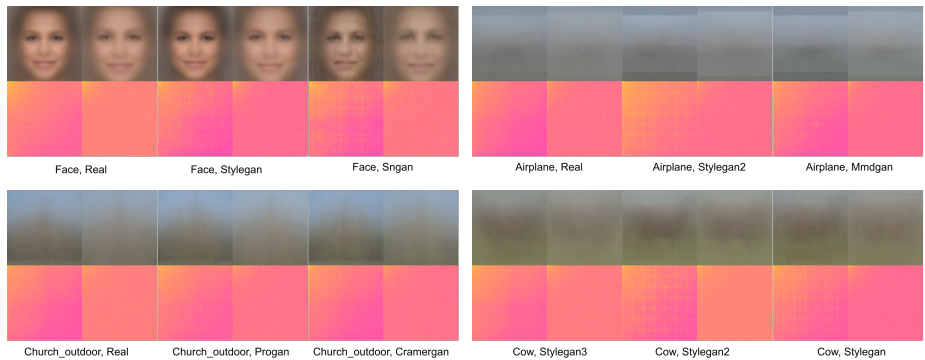

**Fig. 4.** Mean images and spectrums of several generator classes and semantics. For each inset, top: mean image before (left) and after (right) random perturbation, bottom: corresponding mean of DCT images.

**Table 2.** Code repositories of the baseline methods reported in Tab.1 of the main paper.

| Baseline | Codebase |
| --- | --- |
| Yu *et al.* [13], Eigen Face [12] | https://github.com/ningyu1991/GANFingerprints |
| DCT-CNN [4], PRNU [10] | https://github.com/RUB-SysSec/GANDCTAnalysis |
| Reverse Eng. [1] | https://github.com/vishal3477/Reverse_Engineering_GMs |

augmentation strategy (*i.e.* ImageNet-C [5]) for fair comparisons with our proposed method. For Yu *et al.* [13] approach, we also re-implement their method using Pytorch instead of Tensorflow and add an extra 256-D FC layer and report the improved performance in Tab.1 of the main paper. For DCT-CNN [4], we observe that the training of their recommended architecture, a 1-layer multinomial regression model, collapses due to the complexity of Attribution88 benchmark. We therefore use a Resnet50 backbone (same as RepMix) instead.

## 3 Further results

### 3.1 More backbones

Tab. 3 shows performance of RepMix on more advanced backbones, Resnet101 and Densenet121, on the ablated set (c.f. Tab.3 of the main paper). We achieve slightly better performance at the cost of more expensive models.

### 3.2 Beta distribution in the RepMix layer

Fig. 5 shows our experiment with the beta distribution in RepMix layer on the ablated set of Attribution88 benchmark. RepMix is robust to $\beta$ value of 0.25

Table 3. RepMix with Resnet101 and Densenet121 backbones on the ablated set.

| Backbone | Detection Acc. ⇑ | Attribution Acc. ⇑ | Attribution NMI ⇑ |
|---|---|---|---|
| Resnet101 | **0.9523** | 0.7430 | 0.5546 |
| Densenet121 | 0.9504 | **0.7474** | **0.5673** |

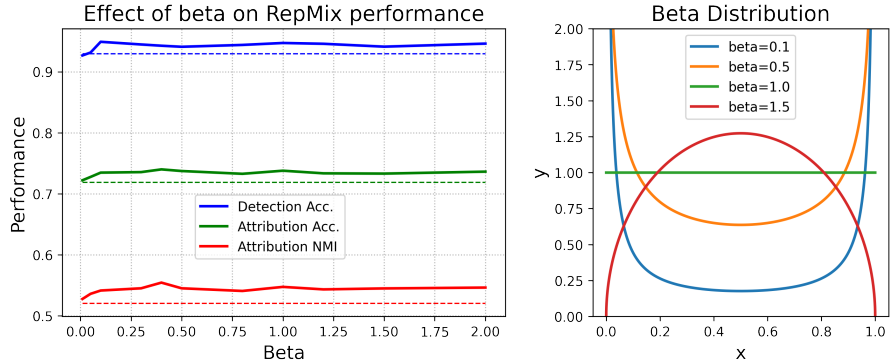

**Fig. 5.** Effects of beta on RepMix. Dash lines refer to the no-mixing baselines.

and above, performing favorably even at uniform distribution ($\beta = 1.0$). We argue that the RepMix layer is positioned close to the loss layer in our network, resulting in a stronger regularization and less dependence on the value of the mixing ratio.

### 3.3  Confusion matrix

Fig. 6 shows in details the contribution of each source class's performance towards the overall attribution accuracy of RepMix and other baselines. It is consistent across all deep-learning methods that attribution performance is high for Real, Progan and SNgan, probably because of the unique data distribution (Real) or distinct GAN architecture (Progan, SNgan). The source of confusion mainly comes from attributing images from the Stylegan family and Cramergan/MMDgan because of the similarities in design of these GAN architectures.

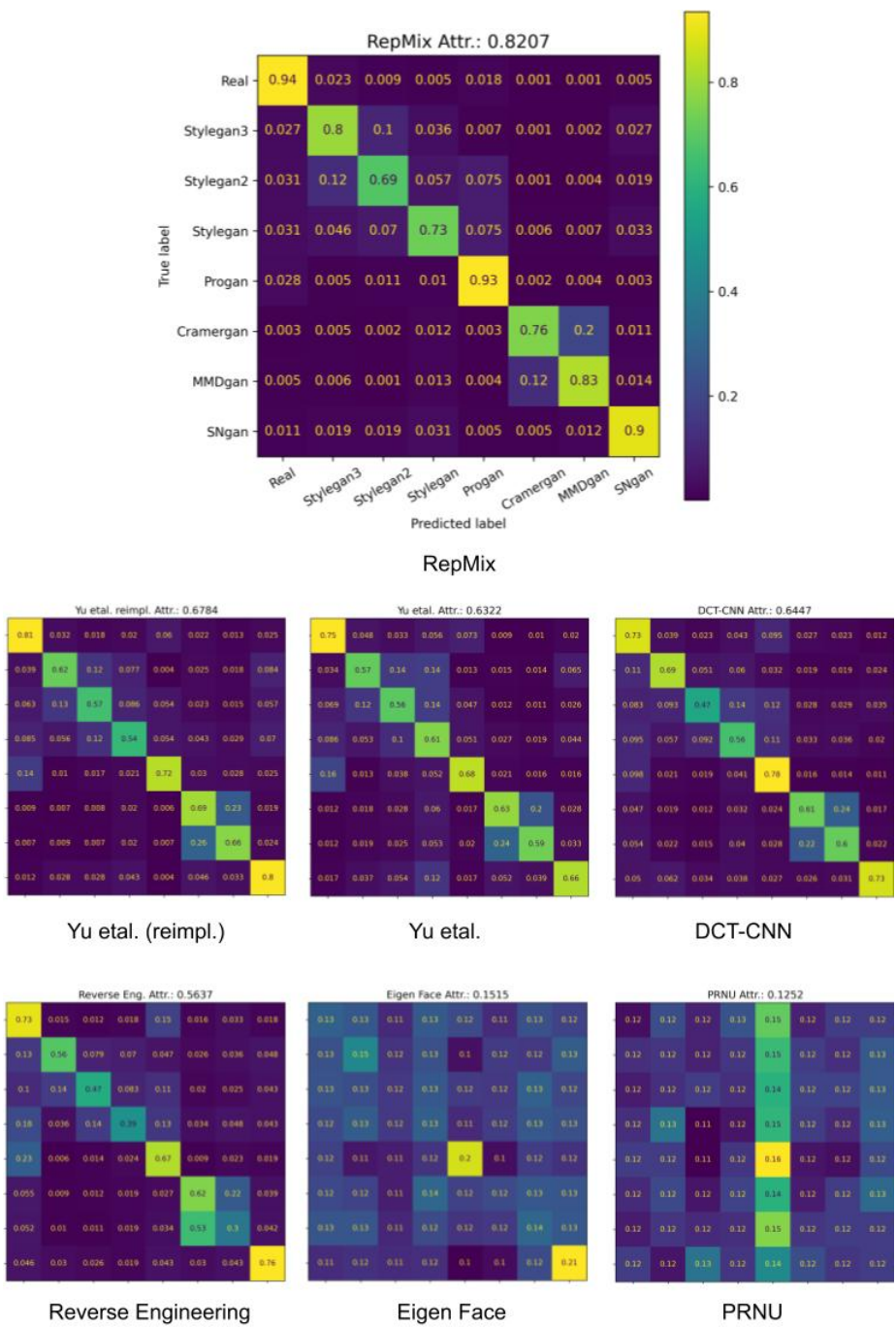

**Fig. 6.** Confusion matrices of RepMix and all baselines on Attribution88.

8        T. Bui et al....



\end{document}